\documentclass[runningheads]{llncs}
\usepackage{graphicx}
\usepackage{amsfonts}
\begin{document}
\title{X2Teeth: 3D Teeth Reconstruction from a Single Panoramic Radiograph}
%
%
\author{
    Yuan Liang \and
    Weinan Song \and
    Jiawei Yang  \and
    Liang Qiu  \and
    Kun Wang  \and 
    Lei He
    }
\institute{
    University of California, Los Angeles, CA 90095, USA \\
    \email{lhe@ee.ucla.edu}
    }
\maketitle              
\begin{abstract}
3D teeth reconstruction from X-ray is important for dental diagnosis and many clinical operations. 
However, no existing work has explored the reconstruction of teeth for a whole cavity from a single panoramic radiograph. 
Different from single object reconstruction from photos, this task has the unique challenge of constructing multiple objects at high resolutions.
To conquer this task, we develop a novel ConvNet \textit{X2Teeth} that decomposes the task into teeth localization and single-shape estimation. 
We also introduce a patch-based training strategy, such that \textit{X2Teeth} can be end-to-end trained for optimal performance. 
Extensive experiments show that our method can successfully estimate the 3D structure of the cavity and reflect the details for each tooth. 
Moreover, \textit{X2Teeth} achieves a reconstruction IoU of 0.681, which significantly outperforms the encoder-decoder method by $1.71\times$ and the retrieval-based method by $1.52\times$.
Our method can also be promising for other multi-anatomy 3D reconstruction tasks. 

\end{abstract}
\section{Introduction}
X-ray is an important clinical imaging modality for dental diagnosis and surgical operations. 
Compared to Cone Beam Computed Tomography (CBCT), X-ray exceeds in lower cost and less absorbed dose of radiation. 
However, X-ray imagery cannot provide 3D information about teeth volumes or their spatial localization, which is useful for many dental applications \cite{buchaillard2004reconstruction,rahimi20053d}, \textit{e.g.}, micro-screws planning, root alignment assessment, and treatment simulations. 
Moreover, with volumetric radiation transport, the understanding and interpretation of X-ray imagery can be only apparent to experienced experts \cite{henzler2018single}. 
As such, the 3D visualization of X-rays can also be beneficial for applications related to patient education and physician training. 

There have been several researches on the 3D reconstruction of a single tooth from its 2D scanning. 
For example, \cite{mazzotta20132d} models the volume of a tooth from X-rays by deforming the corresponding tooth atlas according to landmark aligning.
\cite{abdelrahim2012realistic,abdelrehim20132d} reconstruct a tooth from its crown photo by utilizing the surface reflectance model with shape priors.   
Despite those work, no one has explored the 3D teeth reconstruction of a whole cavity from a single panoramic radiograph. 
This task is more challenging than the single tooth reconstruction, since not only tooth shapes but also spatial localization of teeth should be estimated from their 2D representation. 
Moreover, all the existing methods of tooth reconstruction \cite{abdelrahim2012realistic,abdelrehim20132d,mazzotta20132d} utilize ad-hoc image processing steps and handcrafted shape features. 
Currently, Convolutional Neural Networks (ConvNet) provide an accurate solution for single-view 3D reconstruction by discriminative learning, and have become the state-of-the-art for many photo-based benchmarks \cite{choy20163d,henzler2018single,tatarchenko2017octree}. 
However, the application of ConvNet on the teeth reconstruction has not yet been explored. 

In this work, we pioneer the study of 3D teeth reconstruction of the whole cavity from a single panoramic radiograph with ConvNet. 
Different from most 3D reconstruction benchmarks \cite{chang2015shapenet,sun2018pix3d}, which target at estimating a single volume per low-resolution photo, our task has the unique challenge to estimate the shapes and localization of multiple objects at high resolutions. 
As such, we propose \textit{X2Teeth}, an end-to-end trainable ConvNet that is compact for multi-object 3D reconstruction. 
Specifically, \textit{X2Teeth} decomposes the reconstruction of teeth for a whole cavity into two sub-tasks of teeth localization and patch-wise tooth reconstruction. 
Moreover, we employ the random sampling of tooth patches during training guided by teeth localization to reduce the computational cost, which enables the end-to-end optimization of the whole network. 
According to experiments, our method can successfully reconstruct the 3D structure of the cavity, as well as restore the teeth with details at high resolutions. 
Moreover, we show \textit{X2Teeth} achieves the reconstruction Intersection over Union (IoU) of 0.6817, outperforming the state-of-the-art encoder-decoder method by $1.71\times$ and retrieval-based method by $1.52\times$, which demonstrates the effectiveness of our method. 
To the best of our knowledge, this is the first work that explores 3D teeth reconstruction of the whole cavity from a single panoramic radiograph. 

\begin{figure}
\includegraphics[width=\textwidth]{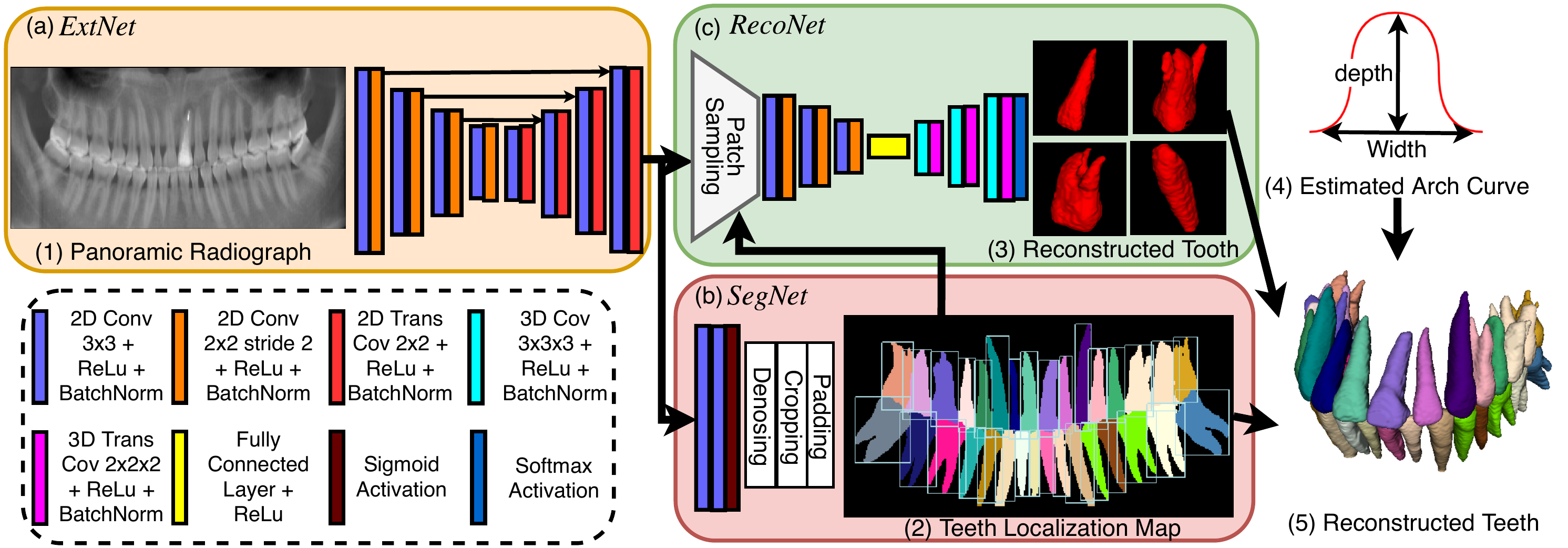}
\caption{Overall architecture of \textit{X2Teeth}. \textit{X2Teeth} consists of three subnets: (a) \textit{ExtNet}, (b) \textit{SegNet} and (c) \textit{ReconNet}. \textit{ExtNet} captures deep representations of teeth from the input panoramic radiograph. Based on the representations, \textit{SegNet} performs pixel-wise classification followed by segmentation map denoising for localizing teeth. \textit{ReconNet} samples tooth patches from the derived feature map and performs single-shape reconstruction. The final reconstruction of the whole cavity is the assembling of the reconstructed teeth according to the teeth localization and arch curve that estimated via $\beta$ function model. The whole model can be end-to-end trained.} \label{fig1}
\end{figure}

\section{Methodologies}
Fig. \ref{fig1} shows the overall architecture of our \textit{X2Teeth}. 
We define the input of \textit{X2Teeth} as a 2D panoramic radiograph (Fig.\ref{fig1}(1)), and the output as a 3D occupancy grid (Fig.\ref{fig1}(5)) of multiple categories for indicating different teeth. 
Different from the existing single-shape estimations \cite{choy20163d,tatarchenko2017octree} that mostly employ a single encoder-decoder structure for mapping the input image to one reconstructed object, \textit{X2Teeth} decomposes the task into object localization (Fig.\ref{fig1}(b)) and patch-wise single tooth reconstruction (Fig.\ref{fig1}(c)).
As such, the reconstruction can be carried out at high resolutions for giving more 3D details under the computational constraint, since tensor dimensions within the network can be largely reduced compared to directly reconstructing the whole cavity volume. 
Moreover, both sub-tasks share a feature extraction subnet (Fig.\ref{fig1}(a)), and the whole model can be end-to-end optimized  by employing a sampling-based training strategy for the optimal performance. 
With the derived teeth localization and tooth volumes, the final reconstruction of the cavity is derived by assembling different objects along the dental arch that is estimated via a $\beta$ function model.  

\subsection{Model Architecture}
Given the panoramic radiograph, our \textit{X2Teeth} consists of three components: (1) a feature extracting subnet \textit{ExtNet} for capturing teeth representations, (2) a segmentation subnet \textit{SegNet} for estimating teeth localization, and (3) a patch-wise reconstruction subnet \textit{ReconNet} for estimating the volume of a single tooth from the corresponding feature map patch.
The detailed model configuration can be seen from the Fig.\ref{fig1}.

\subsubsection{\textit{ExtNet}} As shown in Fig.\ref{fig1}(a), \textit{ExtNet} has an encoder-decoder structure consisting of 2D convolutions for capturing contexture features from the input panoramic radiograph (Fig.\ref{fig1}(1)).
The extracted features are at high resolutions as the input image, and are trained to be discriminative for both \textit{SegNet} and \textit{ReconNet} to increase the compactness of the network. 
\textit{ExtNet} utilizes strided convolutions for down-sampling and transpose convolutions for up-sampling, as well as channel concatenations between different layers for feature fusion.   

\subsubsection{\textit{SegNet}} Given the feature map of \textit{ExtNet}, \textit{SegNet} maps it into a categorical mask $Y_{seg} \in \mathbb{Z}^{H \times W \times C}$, where $H$ and $W$ are image height and width, while $C$ denotes the number of categories of teeth. 
Especially, a categorical vector $y \in Y_{mask}$ is multi-hot encoded, since nearby teeth can overlap in a panoramic radiograph because of the 2D projecting. 
With the categorical mask, \textit{SegNet} further performs denoising by keeping the largest island of segmentation per tooth type, and localizes teeth by deriving their bounding boxes as shown in Fig.\ref{fig1}(2). 
As indicated in Fig.\ref{fig1}(b), \textit{SegNet} consists of 2D convolutional layers followed by a \textit{Sigmoid} transfer in order to perform the multi-label prediction. 
In our experiments, we set $C=32$ for modeling the full set of teeth of an adult, including the four wisdom teeth that possibly exist for some individuals. 

\subsubsection{\textit{ReconNet}} \textit{ReconNet} samples the feature patch of a tooth, and maps the 2D patch into the 3D occupancy probability map $Y_{recon} \in \mathbb{R}^{H_{p} \times W_{p} \times D_{p} \times 2}$ of that tooth, where $H_{p}$, $W_{p}$, $D_{p}$ are patch height, width and depth, respectively. 
The 2D feature patch is cropped from the feature map derived from \textit{ExtNet}, while the cropping is guided by the tooth localization derived from \textit{SegNet}. 
Similar to \cite{choy20163d,henzler2018single}, \textit{ReconNet} has an encoder-decoder structure consisting of both 2D and 3D convolutions. 
The encoder employs 2D convolutions, and its output is flattened into a 1D feature vector for the fully connected operation; while the decoder employs 3D convolutions to map this feature vector into the target dimension.
Since our \textit{ReconNet} operates on small image patches rather than the whole x-ray, the reconstruction can be done at high resolutions for restoring the details of teeth. 
In this work, we set $H_{p}=120$, $W_{p}=60$, $D_{p}=60$ since all teeth fit into this dimension. 

\subsubsection{\textit{Teeth Assembling}} By assembling the predicted tooth volumes according to their estimated localization from x-ray segmentation, we can achieve the 3D reconstruction of the cavity as a flat plane without the depth information about the cavity. 
This reconstruction is an estimation for the real cavity that is projected along the dental arch. 
Many previous work has investigated the modeling and prediction of the dental arch curve \cite{noroozi2001dental}. 
In this work, we employ the $\beta$ function model introduced by \cite{braun1998form}, which estimates the curve by fitting the measurements of cavity depth and width (Fig.\ref{fig1}(4)). 
As the final step, our prediction of teeth for the whole cavity (Fig.\ref{fig1}(5)) can be simply achieved by bending the assembled flat reconstruction along the estimated arch curve. 

\subsection{Training Strategy}
The loss function of \textit{X2Teeth} is composed of two parts: segmentation loss $L_{seg}$ and patch-wise reconstruction loss $L_{recon}$. 
For $L_{seg}$, considering that a pixel can be of multiple categories because of teeth overlaps on X-rays, we define the segmentation loss as the average of dice loss across all categories. 
Denote the segmentation output $Y_{seg}$ at a pixel $(i,j)$ to be a vector $Y_{seg}(i,j)$ of length $C$, where $C$ is the number of possible categories, then 
\begin{equation}
L_{seg}(Y_{seg}, Y_{gt}) = 1 - \frac{1}{C}\sum_{C}\frac{ \sum_{i,j}{2Y_{seg}(i,j)Y_{gt}{(i,j)}}}{\sum_{i,j}{(Y_{seg}(i,j) + Y_{gt}(i,j))}}, 
\end{equation}
where $Y_{gt}$ is the multi-hot encoded segmentation ground-truth. 
For $L_{recon}$, we employ the 3D dice loss for defining the difference between the target and the predicted volumes.
Let the reconstruction output $Y_{recon}$ at a pixel $(i,j,k)$ be a Bernoulli distribution $Y_{recon}(i,j,k)$, then 
\begin{equation}
L_{recon}(Y_{recon}, Y_{gt}) = 1 - 2\frac{ \sum_{c=1}^{2}\sum_{i,j,k}{Y_{recon}(i,j,k)Y_{gt}{(i,j,k)}}}{\sum_{c=1}^{2}\sum_{i,j,k}{(Y_{recon}(i,j,k) + Y_{gt}(i,j,k)})}, 
\end{equation}
where $Y_{gt}$ is the reconstruction ground-truth. 

We employ a two-stage training paradigm. 
In the first stage, we train \textit{ExtNet} and \textit{SegNet} for the teeth localization by optimizing $L_{seg}$, such that the model can achieve an acceptable tooth patch sampling accuracy. In the second stage, we train the whole \textit{X2Teeth} including \textit{ReconNet} by optimizing the loss sum $L=L_{seg}+L_{recon}$ for both localization and reconstruction.
Note that Adam optimizer is used for optimization.
For each GPU, we set the batch size of panoramic radiograph as 1, and the batch size of tooth patches as 10. 
Besides, standard augmentations are employed for images, including random shifting, scaling, rotating and adding Gaussian noise. 
Finally, we implement our framework in Pytorch, and trained for the experiments on three NVidia Titan Xp GPUs. 

\section{Experiments}
In this section, we validate and demonstrate the capability of our method for the teeth reconstruction from the panoramic radiograph. 
First, we introduce our in-house dataset of X-ray and panoramic radiograph pairs with teeth annotations from experts.  
Second, we validate \textit{X2Teeth} by comparing with two state-of-the-art single view 3D reconstruction methods. 
Finally, we look into the performance of \textit{X2Teeth} on the two sub-tasks of teeth localization and single tooth reconstruction. 

\subsection{Dataset}
Ideally, we need paired data of the panoramic radiographs and CBCT scans captured from the same subject to train and validate \textit{X2Teeth}.
However, in order to control the radiation absorbed by subjects, such data pairs can rarely be collected in clinical settings. 
Therefore, we take an alternative approach by collecting high resolution CBCT scans and synthesize their corresponding panoramic radiographs. 
Such synthesis is valid since CBCT scans contain full 3D information of cavity, while panoramic radiographs are the 2D projections of them. 
Several previous work has demonstrated promising results for high quality synthesis, and in our work, we employ the method of Yun \textit{et al.} \cite{yun2019automatic} for building our dataset. 
Our in-house dataset contains 23 pairs of 3D CBCT scans and panoramic radiographs, each with a resolution ranging form 0.250 mm to 0.434 mm. 
All CBCT scans and panoramic radiographs are first labeled with pixel-wise tooth masks by 3 annotators, and then reviewed by 2 board-certificated dentists. 
Finally, we randomly split the dataset into 15 pairs for training, 1 pair for validation, and 7 pairs for testing.  

\subsection{Overall Evaluation of Teeth Reconstruction}
\begin{table}
\centering
\caption{Comparison of reconstruction accuracy between \textit{X2Teeth} and general purpose reconstruction methods in terms of IoU, detection accuracy (DA) and identification accuracy (FA). We report each metric in the format of \textit{mean} $\pm$ \textit{std}.}\label{tab1}
\begin{tabular}{l | c c c}
\hline
Method &  IoU & DA & DF \\
\hline
3D-R2N2 & 0.398 $\pm$ 0.183 & 0.498 $\pm$ 0.101 & 0.592 $\pm$ 0.257 \\
DeepRetrieval & 0.448 $\pm$ 0.116 & 0.594 $\pm$ 0.088 & 0.503 $\pm$ 0.119 \\
\textbf{X2Teeth (ours)} & \textbf{0.682 $\pm$ 0.030} & \textbf{0.702 $\pm$ 0.042} & \textbf{0.747 $\pm$ 0.038}\\
\hline
\end{tabular}
\end{table}
\begin{figure}
\includegraphics[width=\textwidth]{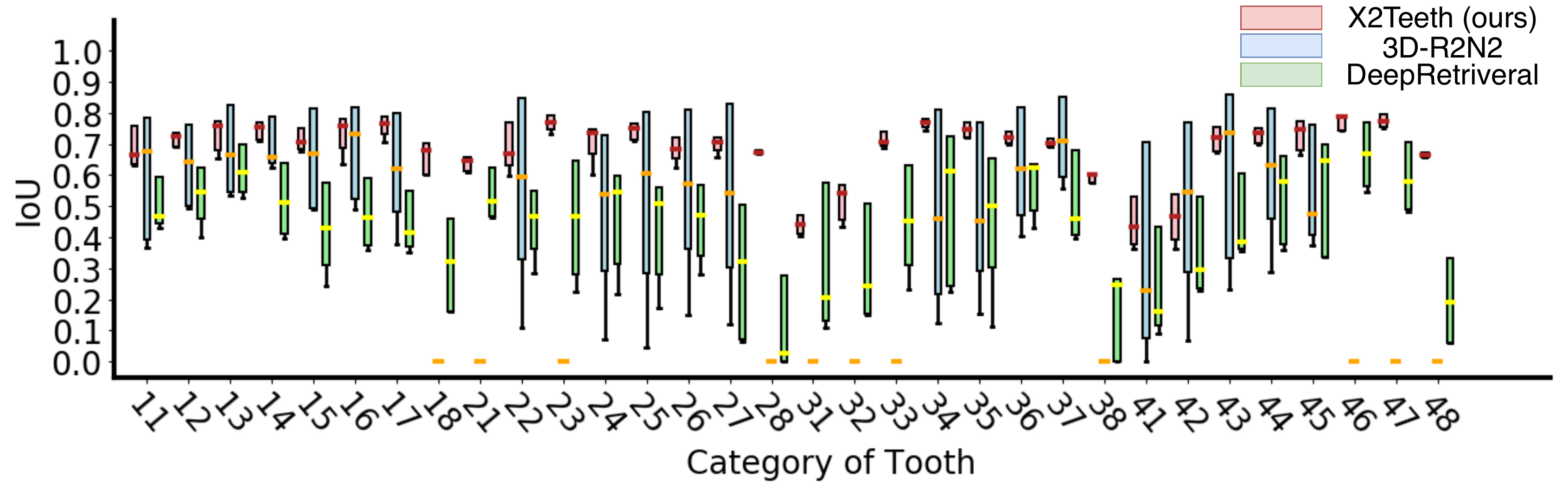}
\caption{IoU comparison of different tooth types between \textit{X2Teeth}, 3D-R2N2, and DeepRetrieval.} \label{fig2}
\end{figure}

We compare our \textit{X2Teeth} with two general purpose reconstruction methods that have achieved state-of-the-art performance as baselines: 3D-R2N2 \cite{choy20163d} and DeepRetrieval \cite{tatarchenko2019single}. 
3D-R2N2 employs an encoder-decoder network to map the input image to a latent representation, and reasons about the 3D structure upon it.
To adapt 3D-R2N2 for high resolution X-rays in our tasks, we follow \cite{tatarchenko2019single} by designing the output of the model to be $128^3$ voxel grids, and up-sampling the prediction to the original resolution for evaluation. 
DeepRetrieval is a retrieval-based method that reconstructs images by deep feature recognition.  
Specifically, 2D images are embedded into a discriminative  descriptor by using a ConvNet \cite{krizhevsky2012imagenet} as its representation.
The corresponding 3D shape of a known image that shares the smallest Euclidean distance with the query image according to the representation is then retrieved as the prediction. 

\textbf{Quantitative Comparison.} We evaluate the performance of models with intersection over union (IoU) between the predicted and the ground-truth voxels, as well as detection accuracy (DA) and identification accuracy (FA) \cite{cui2019toothnet}.  
The formulations of the metrics are:
\begin{equation}
IoU=\frac{|D \cap G|}{|D \cup G|},~~DA=\frac{|D|}{|D \cap G|} ~~and~~FA=\frac{|D \cap G|}{|D|}, 
\end{equation}
where \textit{G} is the set of all teeth in ground-truth data, and \textit{D} is the set of predicted teeth. 
As shown in Table \ref{tab1}, \textit{X2Teeth} outperforms both baseline models significantly in terms of all three metrics.
Specifically, \textit{X2Teeth} achieves a mean IoU of 0.682, which outperforms 3D-R2N2 by $1.71\times$, and DeepRetrieval $1.52\times$.  
Similarly, Fig.\ref{fig2} reveals IoUs for all the 32 types of tooth among the three methods, where our method has the highest median and the smallest likely range of variation (IQR) for all tooth types, which shows the consistent accuracy of \textit{X2Teeth}.
Yet, we also find that all algorithms have a lower accuracy for wisdom teeth (numbering 18, 28, 38, and 48) than the other teeth, indicating that the wisdom teeth are more subject-dependent, and thus difficult to predict. 

\begin{figure} [t!]
\centering
\includegraphics[width=\textwidth]{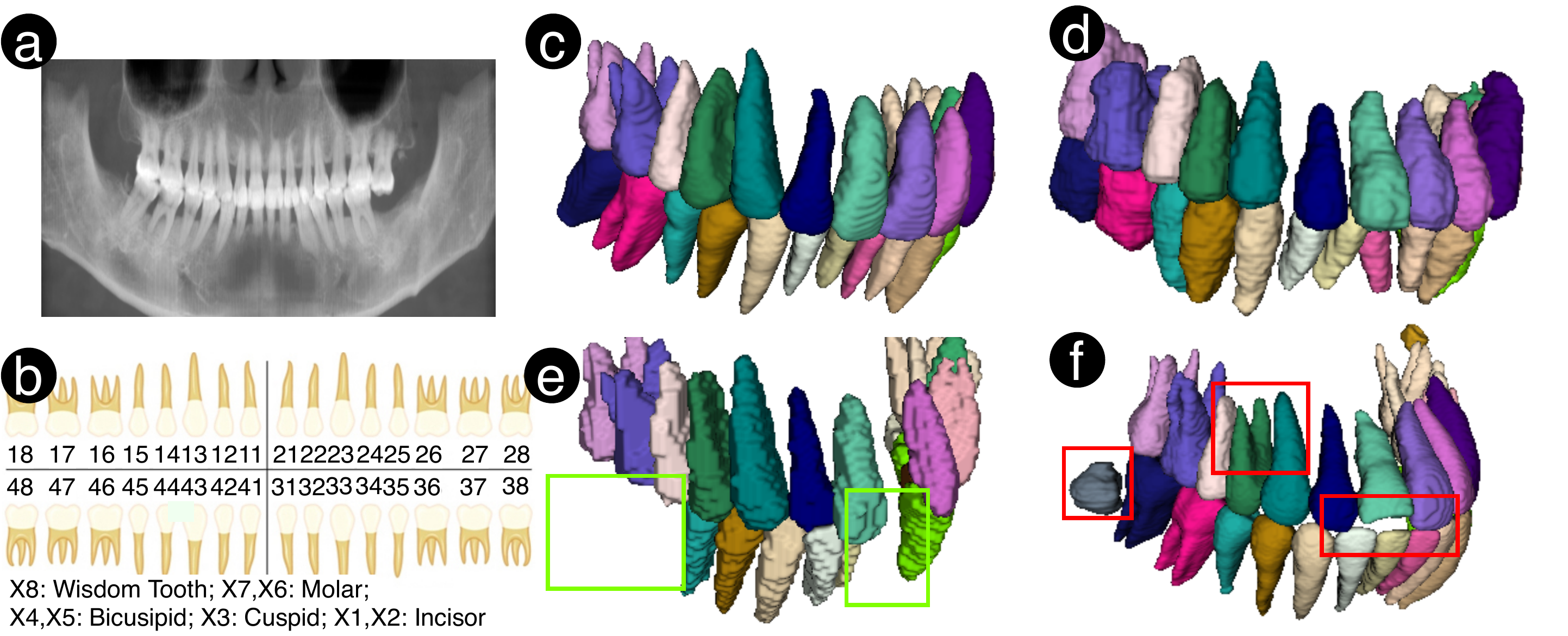}
\caption{Comparison of the reconstruction between (d) \textit{X2Teeth} (ours), (e) 3D-R2N2, and (f) DeepRetrieval. (a) shows the input panoramic radiograph from the testing set, (c) shows the ground-truth of reconstruction, and (b) is the teeth numbering rule.} 
\label{fig3}
\end{figure}
\textbf{Qualitative Comparison.} Fig.\ref{fig3} visualizes the 3D reconstructions of a panoramic radiograph (Fig.\ref{fig3}(a)) from the testing set, which clearly shows our \textit{X2Teeth} can achieve more appealing results than the other two methods. 
As for 3D-R2N2, its reconstruction (Fig.\ref{fig3}(e)) misses several teeth in the prediction as circled with green boxes, possibly because spatially small teeth can lose their representations within the deep feature map during the deep encoding process. 
The similar issue of missing tooth in predictions has also been previously reported in some teeth segmentation work \cite{cui2019toothnet}. 
Moreover, the reconstruction of 3D-R2N2 has coarse object surfaces that lack details about each tooth. 
This is because 3D-R2N2 is not compact enough and can only operate at the compressed resolution. 
As for DeepRetrieval, although the construction (Fig.\ref{fig3}(f)) has adequate details of teeth since its retrieved from high-resolution dataset, it fails to reflect the unique structure of individual cavity.  
The red boxes in Fig.\ref{fig3}(f) point out the significant differences in wisdom teeth, tooth root shapes, and teeth occlusion between the retrieved teeth and the ground-truth. 
Comparing to these two methods, \textit{X2Teeth} has achieved a reconstruction (Fig.\ref{fig3}(d)) that can reflects both the unique structure of cavity and the details of each tooth, by formulating the task as the optimization of two sub-tasks for teeth localization and single tooth reconstruction.  

\subsection{Sub-task Evaluations}
\begin{figure} [ht!]
\centering
\includegraphics[width=0.85\textwidth]{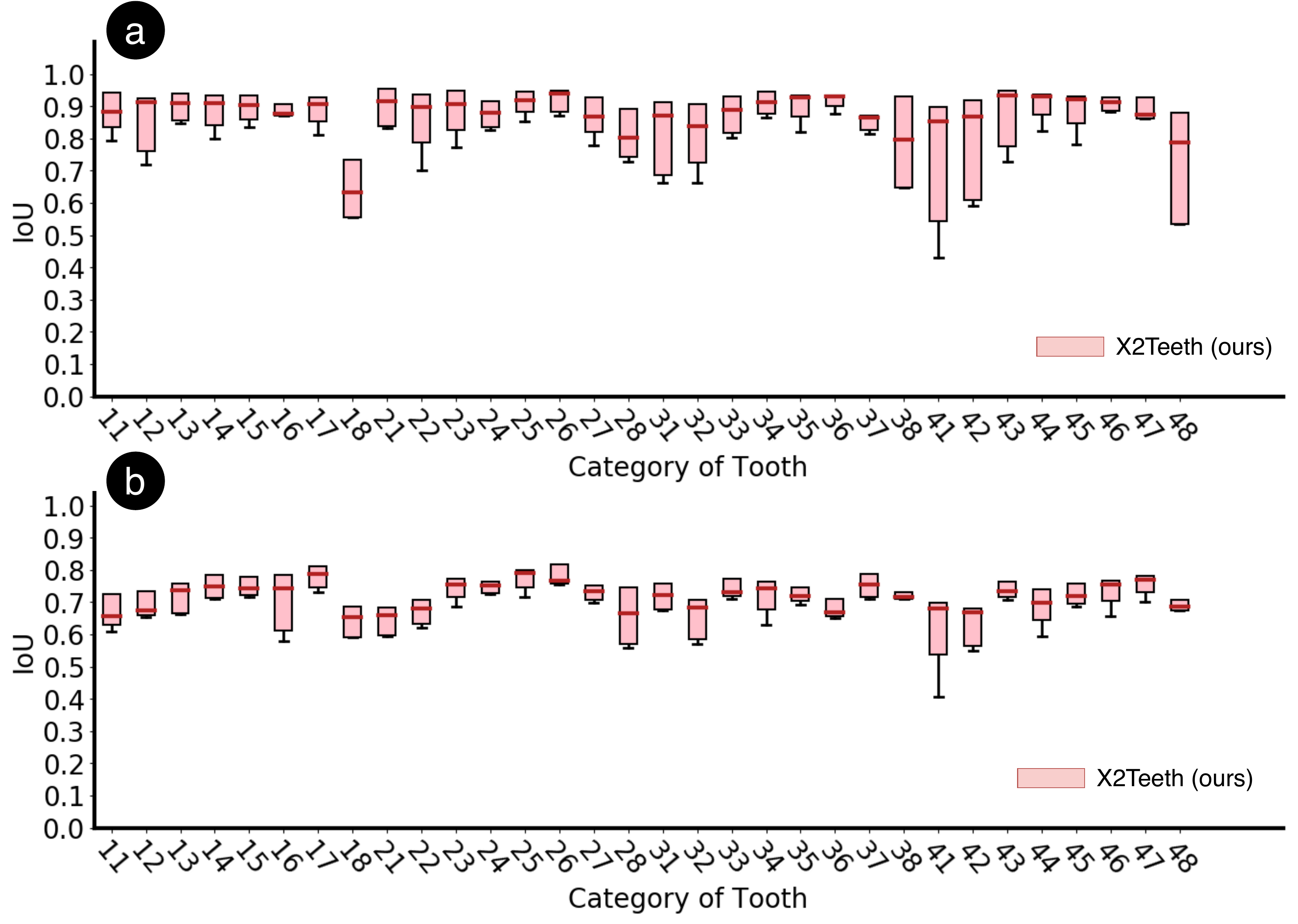}
\caption{(a) Segmentation IoUs of various teeth for the teeth localization sub-task. (b) Reconstruction IoUs of various teeth for the single tooth reconstruction sub-task.} \label{fig4}
\end{figure}
For better understanding the performance of \textit{X2Teeth}, we evaluate its accuracy on the two sub-tasks of teeth localization and single tooth reconstruction. 
Fig.\ref{fig4}(a) shows the IoUs of different teeth for the 2D segmentation, where our method achieves an average IoU of 0.847$\pm$0.071. 
The results validate that \textit{X2Teeth} can accurately localize teeth, which enables the further sampling of tooth patches for the patch-based reconstruction. 
We also observe that the mean segmentation IoU for the 4 wisdom teeth (numbering X8) is 0.705$\pm$0.056, which is lower than the other teeth. 
This is possibly because they have lower contrasts with surrounded bone structures, such that are more challenging to segment. 
Fig. \ref{fig4}(b) demonstrates the IoUs of different types of teeth for the single tooth reconstruction, where our method achieves a mean IoU of 0.707$\pm$0.044. 
Still, wisdom teeth have the significantly lower mean IoU of 0.668$\pm$0.050, which can be contributed by the lower contrast with surroundings, less accurate localization, and the subject-dependent nature of their shapes. 
Moreover, incisor teeth (numbering X1 and X2) are observed to have less accurate reconstructions with the mean IoU of 0.661$\pm$0.031.
We argue the reason can be their feature vanishing in the deep feature maps considering their small spatial size. 
\section{Conclusion}
In this paper, we initialize the study of 3D teeth reconstruction of the whole cavity from a single panoramic radiograph. 
In order to solve the challenges posed by the high resolution of images and multi-object reconstruction, we propose \textit{X2Teeth} to decompose the task into teeth localization and single tooth reconstruction. 
Our \textit{X2Teeth} is compact and employs sampling-based training strategy, which enables the end-to-end optimization of the whole model. 
Our experiments qualitatively and quantitatively demonstrate that \textit{X2Teeth} achieves accurate reconstruction with tooth details. 
Moreover, our method can also be promising for other multi-anatomy 3D reconstruction tasks. 

%
%

\begin{thebibliography}{10}
\providecommand{\url}[1]{\texttt{#1}}
\providecommand{\urlprefix}{URL }
\providecommand{\doi}[1]{https://doi.org/#1}

\bibitem{abdelrahim2012realistic}
Abdelrahim, A.S., El-Melegy, M.T., Farag, A.A.: Realistic 3d reconstruction of
  the human teeth using shape from shading with shape priors. In: 2012 IEEE
  Computer Society Conference on Computer Vision and Pattern Recognition
  Workshops. pp. 64--69. IEEE (2012)

\bibitem{abdelrehim20132d}
Abdelrehim, A.S., Farag, A.A., Shalaby, A.M., El-Melegy, M.T.: 2d-pca shape
  models: Application to 3d reconstruction of the human teeth from a single
  image. In: International MICCAI Workshop on Medical Computer Vision. pp.
  44--52. Springer (2013)

\bibitem{braun1998form}
Braun, S., Hnat, W.P., Fender, D.E., Legan, H.L.: The form of the human dental
  arch. The Angle Orthodontist  \textbf{68}(1),  29--36 (1998)

\bibitem{buchaillard2004reconstruction}
Buchaillard, S., Ong, S.H., Payan, Y., Foong, K.W.: Reconstruction of 3d tooth
  images. In: 2004 International Conference on Image Processing, 2004. ICIP'04.
  vol.~2, pp. 1077--1080. IEEE (2004)

\bibitem{chang2015shapenet}
Chang, A.X., Funkhouser, T., Guibas, L., Hanrahan, P., Huang, Q., Li, Z.,
  Savarese, S., Savva, M., Song, S., Su, H., et~al.: Shapenet: An
  information-rich 3d model repository. arXiv preprint arXiv:1512.03012  (2015)

\bibitem{choy20163d}
Choy, C.B., Xu, D., Gwak, J., Chen, K., Savarese, S.: 3d-r2n2: A unified
  approach for single and multi-view 3d object reconstruction. In: European
  conference on computer vision. pp. 628--644. Springer (2016)

\bibitem{cui2019toothnet}
Cui, Z., Li, C., Wang, W.: Toothnet: automatic tooth instance segmentation and
  identification from cone beam ct images. In: Proceedings of the IEEE
  Conference on Computer Vision and Pattern Recognition. pp. 6368--6377 (2019)

\bibitem{henzler2018single}
Henzler, P., Rasche, V., Ropinski, T., Ritschel, T.: Single-image tomography:
  3d volumes from 2d cranial x-rays. In: Computer Graphics Forum. vol.~37, pp.
  377--388. Wiley Online Library (2018)

\bibitem{krizhevsky2012imagenet}
Krizhevsky, A., Sutskever, I., Hinton, G.E.: Imagenet classification with deep
  convolutional neural networks. In: Advances in neural information processing
  systems. pp. 1097--1105 (2012)

\bibitem{mazzotta20132d}
Mazzotta, L., Cozzani, M., Razionale, A., Mutinelli, S., Castaldo, A.,
  Silvestrini-Biavati, A.: From 2d to 3d: Construction of a 3d parametric model
  for detection of dental roots shape and position from a panoramic
  radiograph—a preliminary report. International journal of dentistry
  \textbf{2013} (2013)

\bibitem{noroozi2001dental}
Noroozi, H., Hosseinzadeh~Nik, T., Saeeda, R.: The dental arch form revisited.
  The Angle Orthodontist  \textbf{71}(5),  386--389 (2001)

\bibitem{rahimi20053d}
Rahimi, A., Keilig, L., Bendels, G., Klein, R., Buzug, T.M., Abdelgader, I.,
  Abboud, M., Bourauel, C.: 3d reconstruction of dental specimens from 2d
  histological images and $\mu$ct-scans. Computer Methods in Biomechanics and
  Biomedical Engineering  \textbf{8}(3),  167--176 (2005)

\bibitem{sun2018pix3d}
Sun, X., Wu, J., Zhang, X., Zhang, Z., Zhang, C., Xue, T., Tenenbaum, J.B.,
  Freeman, W.T.: Pix3d: Dataset and methods for single-image 3d shape modeling.
  In: Proceedings of the IEEE Conference on Computer Vision and Pattern
  Recognition. pp. 2974--2983 (2018)

\bibitem{tatarchenko2017octree}
Tatarchenko, M., Dosovitskiy, A., Brox, T.: Octree generating networks:
  Efficient convolutional architectures for high-resolution 3d outputs. In:
  Proceedings of the IEEE International Conference on Computer Vision. pp.
  2088--2096 (2017)

\bibitem{tatarchenko2019single}
Tatarchenko, M., Richter, S.R., Ranftl, R., Li, Z., Koltun, V., Brox, T.: What
  do single-view 3d reconstruction networks learn? In: Proceedings of the IEEE
  Conference on Computer Vision and Pattern Recognition. pp. 3405--3414 (2019)

\bibitem{yun2019automatic}
Yun, Z., Yang, S., Huang, E., Zhao, L., Yang, W., Feng, Q.: Automatic
  reconstruction method for high-contrast panoramic image from dental cone-beam
  ct data. Computer methods and programs in biomedicine  \textbf{175},
  205--214 (2019)

\end{thebibliography}

\end{document}